\documentclass{article} 
\usepackage{nips15submit_e,times}
\usepackage{hyperref}
\usepackage{url}
\usepackage{textcomp}
\usepackage{graphicx}
\usepackage{amsfonts}
\usepackage{amsmath}
\usepackage{pdfpages}

\newcommand{\ts}{\rule{0pt}{2.6ex}}       
\newcommand{\specialcell}[2][c]{%
  \begin{tabular}[#1]{@{}c@{}}#2\end{tabular}}

\graphicspath{ {figures/}}

\newcommand{\beginsupplement}{%
        \setcounter{table}{0}
        \renewcommand{\thetable}{S\arabic{table}}%
        \setcounter{figure}{0}
        \renewcommand{\thefigure}{S\arabic{figure}}%
     }

\title{Classifying and Segmenting Microscopy Images Using Convolutional Multiple Instance Learning}

\author{
Oren Z.~Kraus,  Lei Jimmy Ba, Brendan Frey \\
University of Toronto \\
\texttt{\{okraus,jimmy,frey\}@psi.utoronto.ca}
}

\nipsfinalcopy 

\begin{document}

\maketitle

\begin{abstract}
Convolutional neural networks (CNN) have achieved state of the art performance on both classification and segmentation tasks. Applying CNNs to microscopy images is challenging due to the lack of datasets labeled at the single cell level. We extend the application of CNNs to microscopy image classification and segmentation using multiple instance learning (MIL). We present the adaptive Noisy-AND MIL pooling function, a new MIL operator that is robust to outliers. Combining CNNs with MIL enables training CNNs using full resolution microscopy images with global labels. We base our approach on the similarity between the aggregation function used in MIL and pooling layers used in CNNs. We show that training MIL CNNs end-to-end outperforms several previous methods on both mammalian and yeast microscopy images without requiring any segmentation steps. 
\end{abstract}

\section{Introduction}

High content screening (HCS) technologies that combine automated fluorescence microscopy with high throughput biotechnology have become powerful systems for studying cell biology and for drug screening \cite{liberali2015single}.
These systems can produce more than 10\textsuperscript{5} images per day, making their success dependent on automated image analysis. Previous analysis pipelines heavily rely on hand-tuning the segmentation, feature extraction, and classification steps for each assay. Although comprehensive tools have become available \cite{eliceiri2012biological}
they are often optimized for mammalian cells and not directly applicable to model organisms such as yeast and \emph{C. elegans}. 
Researchers studying these organisms often manually classify cellular patterns by eye \cite{tkach2012dissecting}. 

Recent advances in deep learning have proven that deep neural networks trained end-to-end can learn powerful feature representations and outperform classifiers built on top of extracted features \cite{krizhevsky2012imagenet,vincent2010stacked}. While object recognition models have been successfully trained using images with one or a few objects of interest at the center of the image, microscopy images often contain hundreds of cells from the label class, as well as a few outliers. Training similar recognition models on HCS screens is therefore challenging due to the lack of datasets labeled at the single cell level.

In this work we describe a convolutional neural network (CNN) that is trained on full resolution microscopy images using multiple instance learning (MIL). The network is designed to produce feature maps for every output category, as proposed for segmentation tasks in \cite{long2014fully}. We pose cellular phenotype classification as a special case of MIL, where each element in a class-specific feature map is considered an instance and each full resolution microscopy image is considered a bag with a label. Typically binary MIL problems assume that a bag is positive if at least one instance within the bag is positive. This assumption does not hold for HCS images due to heterogeneities within cellular populations and imaging artifacts \cite{altschuler2010cellular}. We explore the performance of several global pooling operators on this problem and propose a new operator capable of learning the proportion of instances necessary to activate a label.

The main contributions of our work are the following. We present a unified view of the classical MIL approaches as pooling layers in CNNs and compare their performances. We propose a novel MIL method, ``adaptive Noisy-AND", that is robust to outliers and large numbers of instances. We evaluate our proposed model on both mammalian and yeast datasets, and find that our model significantly outperforms previously published results at phenotype classification. Our model is capable of learning a good classifier for full resolution microscopy images as well as individual cropped cell instances, even though it is only trained using whole image labels. 
We also demonstrate that the model can localize regions with cells in the full resolution microscopy images and that the model predictions are based on activations from these regions.

\section{Related Work}
\label{gen_inst}
\textbf{Current approaches for microscopy image analysis}: Several sophisticated and modular tools \cite{eliceiri2012biological} have been developed for analyzing microscopy images. 
\emph{CellProfiler} \cite{carpenter2006cellprofiler} is a popular tool that was previously used to analyze the datasets described below. All existing tools rely on extracting a large set of predefined features from the original images and subsequently selecting features that are relevant for the learning task. This approach can be limiting for assays that differ from the datasets used to develop these tools. For example, a recent proteome-wide study of protein localization in yeast resorted to evaluating images manually \cite{tkach2012dissecting}. 

Applying deep neural networks to microscopy screens has been challenging due to the lack of large datasets labeled at the single cell level. Other groups have applied deep neural networks to microscopy for segmentation tasks \cite{ning2005toward,ciresan2012deep} using ground truth pixel-level labels. Pachitariu \emph{et al}. \cite{pachitariu2013extracting} use convolutional sparse coding blocks to extract regions of interest from spiking neurons and slices of cortical tissue without supervision. These publications differ from our work as they aim to segment or localize regions of interest within the full resolution images. Here we aim to train a CNN for classifying cellular phenotypes for images of arbitrary size based on only training with weak labels. 

\textbf{Fully convolutional neural networks}: Fully convolutional neural networks (FCNN) have recently achieved state-of-the-art performance on image segmentation tasks \cite{long2014fully,chen2014semantic}. These networks build on the success of networks previously trained on image recognition tasks \cite{krizhevsky2012imagenet,szegedy2014going,simonyan2014very} by converting their fully connected layers to 1x1 convolutions, producing feature maps for each output category. The pre-trained networks are fine-tuned using different techniques to generate output images of the same dimension as input images from the down-sampled feature maps. These networks are trained with pixel level ground truth labels. Pathak \emph{et al}. \cite{pathak2014fully} use MIL with a FCNN to perform segmentation using weak labels. However, dense pixel level ground truth labels are expensive to generate and arbitrary, especially for niche datasets such as microscopy images. In this work we aim to develop a classification CNN using MIL that does not require labels for specific segmented cells, or even require the cells to be segmented. 

\textbf{Multiple instance learning}: Multiple instance learning deals with problems for which labels only exist for sets of data points. In this setting sets of data points are typically referred to as bags and specific data points are referred to as instances. A commonly used assumption for binary labels is that a bag is considered positive if at least one instance within the bag is positive \cite{dietterich1997solving}. Several functions have been used to map the instance space to the bag space. These include Noisy-OR \cite{zhang2005multiple}, log-sum-exponention (LSE) \cite{ramon2000multi}, generalized mean (GM), and the integrated segmentation and recognition (ISR) model \cite{keeler1991integrated}. 
Xu \emph{et al}. \cite{xu2014deep} use the GM pooling function for classifying features extracted from histopathology breast cancer images.

\section{Convolutional MIL Model for Learning Cellular Patterns}

\begin{figure}[h]
\begin{center}
\includegraphics[width=0.75\textwidth]{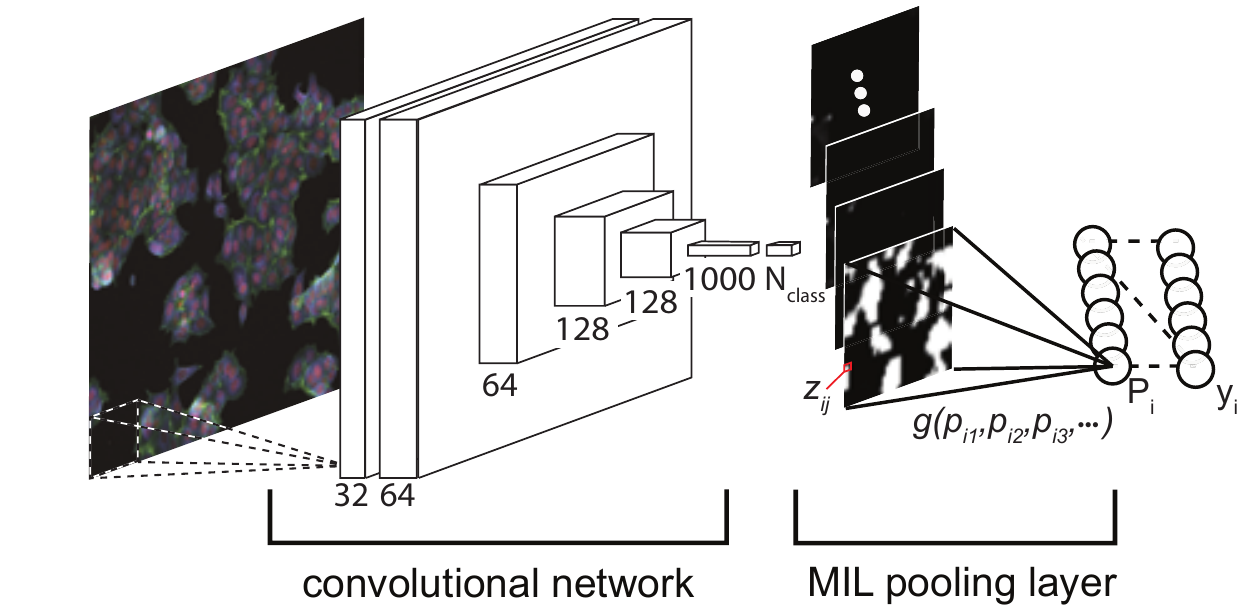}
\end{center}
\vspace{-0.2in}
\caption{Convolutional multiple instance learning model. $g(\cdot)$ is global pooling function that aggregates instance probabilities $p_{ij}$.}
\label{fig:figure_1}
\end{figure}

We propose a CNN capable of classifying microscopy images of arbitrary size, that is trained with only global image level labels. The weakly supervised CNN is designed to output class-specific feature maps representing the probabilities of the classes for different locations in the input image. Our network works as an instance level classifier for individual cropped cells. The same network can also produce an image level classification over images of arbitrary size and varying number of cells through a MIL pooling layer. 

\subsection{Pooling layers as multiple instance learning}
Formally, assuming that the total number of classes is $N_{class}$ for a full resolution image $I$,  we can treat each class $i$ as a separate binary classification problem with label $t_i \in \{0,1\}$. Under the MIL formulation, one is given a bag of $N$ instances that are denoted as $x = \{x_1,\cdots,x_N\}$ and $x_n \in \mathbb{R}^D$ is the feature vector for each instance. The class labels $t_i$ are associated with the entire bag instead of each instance. A binary instance classifier $p(t_i=1|x_j)$ is used to generate predictions $p_{ij}$ across the instances in a bag. The instance predictions $\{p_{ij}\}$ are combined through an aggregate function $g(\cdot)$, e.g. noisy-OR, to map the set of instance predictions to the probability of the final bag label $p(t_i=1 | {x_1, \cdots, x_N})$. In a CNN, each activation in the feature map is computed through the same set of filter weights convolved across the input image. The pooling layers then combine activations of feature maps in convolutional layers. It is easy to see the similarity between the pooling layer and the MIL aggregation function, where features in convolutional layers correspond to instance features $\{x_n\}$ in MIL. In fact, the classical approaches in MIL can be generalized to global pooling layers over the instance bag.  

We formulate the MIL layer in CNNs as a global pooling layer over a class specific feature map for class $i$ referred to as the bag $p_i$. Without loss of generality assume that the $i^{th}$ class specific convolutional layer in a CNN computes a mapping directly from input images to sets of binary instance predictions $I \to \{p_{i1},\cdots,p_{iN}\}$. It first outputs the logit values $z_{ij}$ in the feature map corresponding to instance $j$ in the bag $i$. We define the feature level probability of an instance ${j}$ belonging to class $i$ as $p_{ij}$ where $p_{ij}=\sigma(z_{ij})$ and $\sigma$ is the sigmoid function. The image level class prediction is obtained by applying the global pooling function $g(\cdot)$ over all elements $p_{ij}$. The global pooling function $g(\cdot)$ maps the instance space probabilities to the bag space such that the bag level probability for class $i$ is defined by
\begin{align}
P_i=g(p_{i1},p_{i2},p_{i3},\cdots).
\end{align}
The global pooling function $g(\cdot)$ essentially combines the instance probabilities from each class specific feature map $p_i$ into a single probability. This reduction allows us to train and evaluate the model on inputs of arbitrary size. In the next section we describe the global pooling functions we explored in our experiments. 

While the MIL layer learns the relationship between instances of the same class, the co-occurrence statistics of instances from different classes within the bag could also be informative to identify the classes. We extend our model to learn relationships between classes by adding an additional fully connected layer following the MIL pooling. This layer can either use softmax or sigmoid activations for either multi-class or multi-label problems. We define the softmax output from this layer for each class $i$ as $y_i$. We formulate a joint cross entropy objective function at both the MIL pooling layer and the additional fully connected layer defined by 
\begin{align}
J= -\sum_{i=1}^{N_{class}}\bigg(\log p(t_i|P_i) + \log p(t_i|y_i)\bigg).
\end{align}
$p(t_i|P_i)$ is the binary class prediction from the MIL layer, $p(t_i|P_i)=P_i^{t_i}(1-P_i)^{(1-t_i)}$, and $p(t_i|y_i)$ is either the binary or the multi-class prediction from the fully connected layer.
Our proposed MIL CNN model is shown in figure \ref{fig:figure_1} and is trained using standard error backpropgation. 

\subsection{Global pooling functions}
\label{sec:pooling_functions}
Classifying cellular phenotypes in microscopy images presents a challenging and generalized MIL problem. Due to heterogeneity within cellular populations \cite{altschuler2010cellular} and imaging artifacts, it cannot be assumed that a specific phenotype will not appear in an image not annotated with that label. A more reasonable assumption is that bag labels are determined by a certain proportion of instances being present. Relevant generalizations for MIL have been proposed that assume that all instances collectively contribute to the bag label. 
Here we take an approach similar to \cite{xu2004logistic} in which bag predictions are expressed as the geometric or arithmetic mean of instances, however we adapt the the bag level formulation to model thresholds on instance proportions for different categories.

\begin{figure}[h]
\begin{center}
\includegraphics[width=0.45\textwidth]{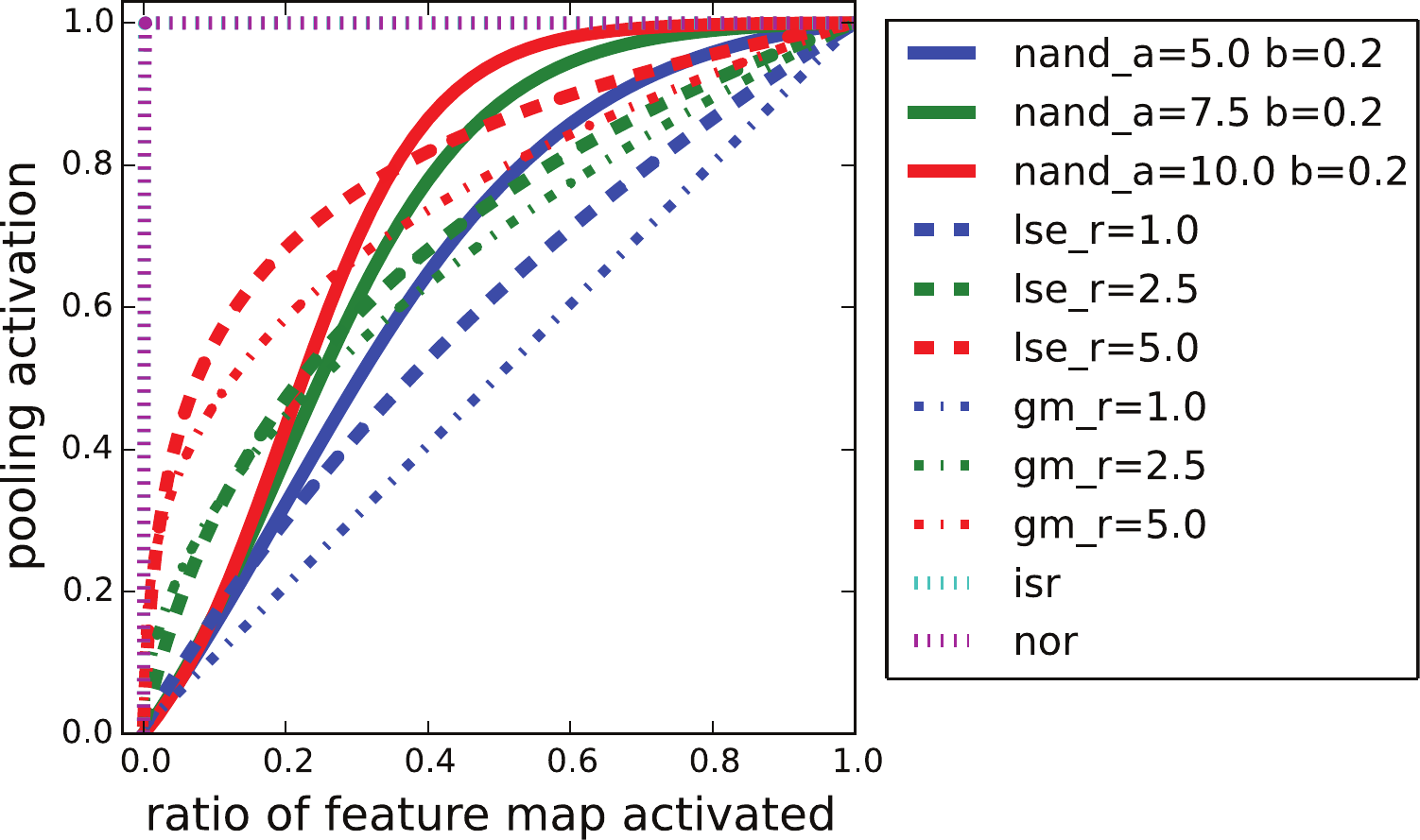} \hspace{0.1in}
\includegraphics[width=0.45\textwidth]{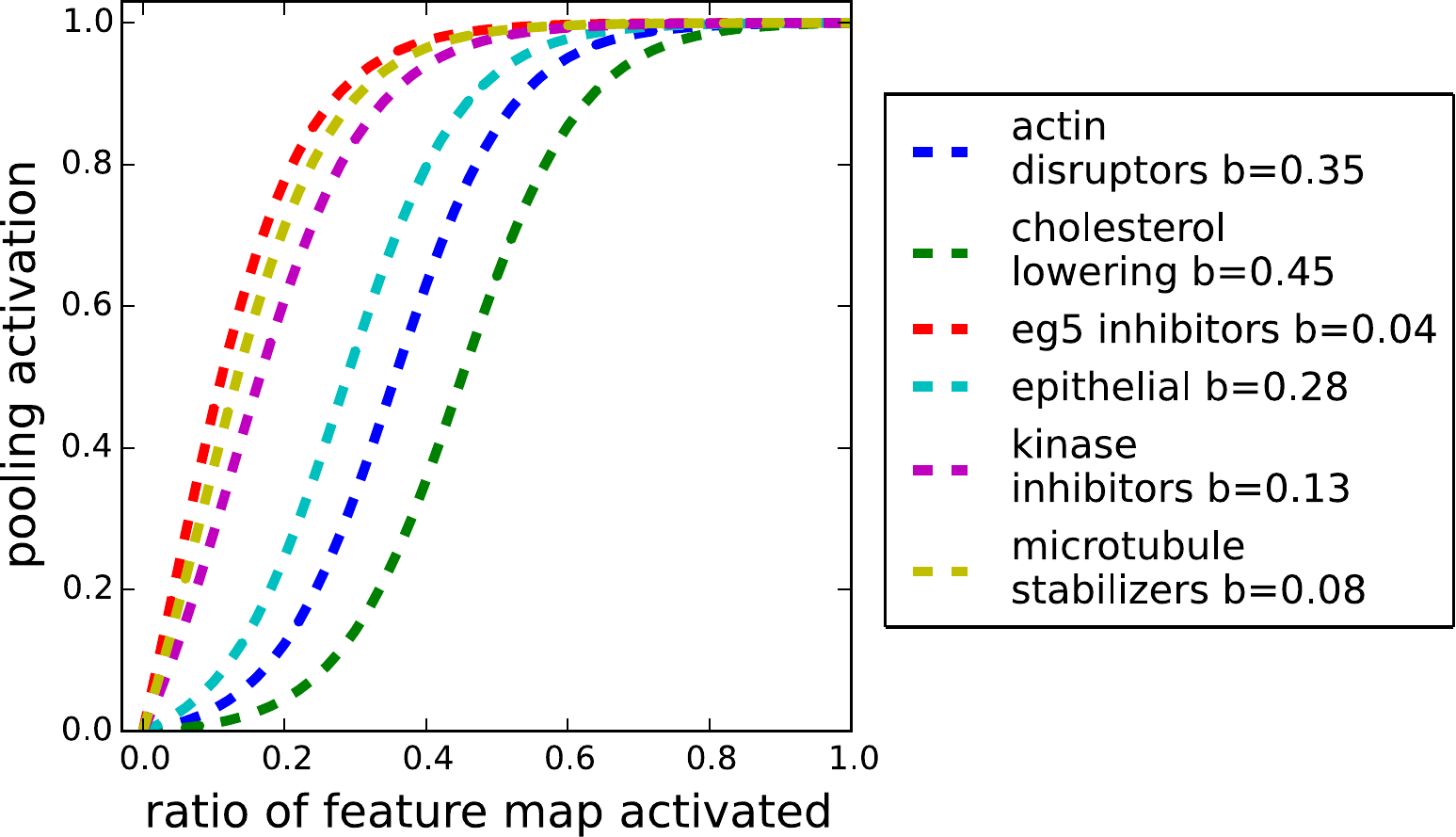}
\end{center}
\vspace{-0.2in}
\caption{MIL Pooling functions. Left, pooling function activations by ratio of feature map activated. Right, activation functions learned by NAND $a_{10}$ for different classes of the breast cancer dataset.}
\label{fig:MIL_pooling_functions}
\end{figure}

We explore the use of several different global pooling functions $g(\cdot)$ in our model. Let $j$ index the instance within a bag. Previously proposed global pooling functions for MIL have been designed as differentiable approximations to the max function in order to satisfy the standard MIL assumption: 
\begin{align}
&g(\{p_{j}\})=1-\prod_{j}(1-p_j) \textrm{\quad  Noisy-OR,} \quad \quad g(\{p_{j}\})=\sum_{j}{p_j\over1 - p_j} \bigg/ \bigg(1 + \sum_{j}{p_j\over1 - p_j}\bigg) \textrm{\quad  ISR,} \nonumber \\
&g(\{p_{j}\})=\bigg({1\over |j|}\sum_j{p_j^r}\bigg)^{1\over r} \textrm{\quad  Generalized mean,} \quad \quad g(\{p_{j}\})={1\over r}\log\bigg({1\over |j|} \sum_j e^{r\cdot p_j}\bigg) \textrm{\quad  LSE.}
\end{align}
We initially attempted to include Noisy-OR \cite{zhang2005multiple} and ISR \cite{keeler1991integrated} in our analysis. We found that both are sensitive to outliers and failed to work with microscopy datasets (as shown in figure~\ref{fig:MIL_pooling_functions}).
LSE and GM both have a parameter $r$ that controls their sharpness. As $r$ increases the functions get closer to representing the max of the instances. In our analysis we use lower values for $r$ than suggested in previous work \cite{ramon2000multi} to allow more instances in the feature maps to contribute to the pooled value. 


\subsubsection{Adaptive Noisy-AND pooling function}
Since existing pooling functions are ill-suited for the task at hand, we examined a different pooling function. Formally, we assume that a bag is positive if the number of positive instances in the bag surpasses a certain threshold. This assumption is meant to model the case of a human expert annotating images of cells by classifying them according to the evident phenotypes or drugs with known targets affecting the majority of the cells in an image. We define the Noisy-AND pooling function as follows,
\begin{align}
P_i=g_i(\{p_{ij}\})=\frac{\sigma(a(\overline{p_{ij}}-b_i)) - \sigma(-ab_i)} {\sigma(a(1-b_i)) - \sigma(-ab_i)} ,\quad  \textrm{where} \quad \overline{p_{ij}}=\frac{1}{|j|}\sum_{j}p_{ij} .
\end{align}
The function 
is designed to activate a bag level probability $P_i$ once the mean of the instance level probabilities $\overline{p_{ij}}$ surpasses a certain threshold. The parameters $a$ and $b_i$ control the shape of the activation function. $b_i$ is a set of parameters learned during training and is meant to represent an adaptable soft threshold for each class $i$. $a$ is fixed parameter that controls the slope of the activation function. 
Figure~\ref{fig:MIL_pooling_functions} shows plots of relevant pooling functions.

\subsection{Localizing cells with Jacobian maps}
\label{sec:localization_method}

\begin{figure}[h]
\begin{center}
\vspace{-0.1in}
\includegraphics[width=0.95\textwidth]{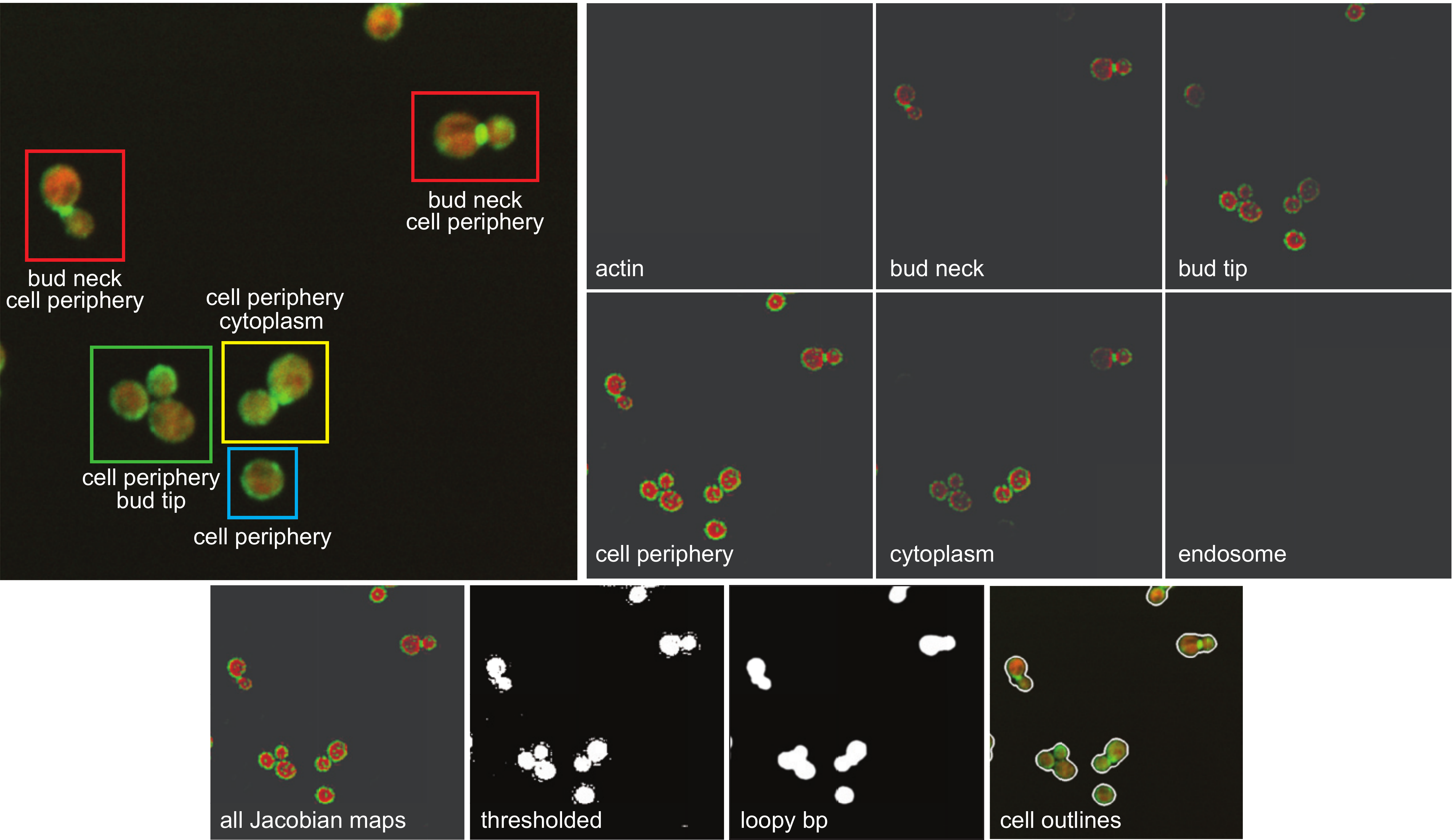}
\end{center}
\vspace{-0.2in}
\caption{Localizing cells with Jacobian maps. Top, yeast cells tagged with a protein that has cell cycle dependent localizations and corresponding Jacobian maps generated from class specific feature maps. Bottom, segmentation by thresholding Jacobian maps and de-noising with loopy bp.}
\label{fig:jacobian_maps}
\end{figure}

Researchers conducting HCS experiments are often interested in obtaining statistics from single cell measurements of their screens. We aimed to extend our model by localizing regions of the full resolution input images that are responsible for activating the class specific feature maps. We employ recently developed methods for visualizing network activations \cite{zeiler2014visualizing,simonyan2013deep} toward this purpose. 
Our approach is similar to Simonyan \emph{et al}. \cite{simonyan2013deep} in which the pre-softmax activations of specific output nodes are back-propagated through a classification network to generate Jacobian maps w.r.t. specific class predictions. Let $a^{(l)}$ be the hidden activations in layer $l$ and $z^{(l)}$ be pre-nonlinearity activations. We define a general recursive non-linear back-propagation process computing a backward activation $\overleftarrow{a}$ for each layer, analogous to the forward propagation:
\begin{align}
\overleftarrow{a}^{(l-1)} = f\bigg({\partial z^{(l)} \over \partial z^{(l-1)}} \overleftarrow{a}^{(l)}\bigg) \quad \textrm{where}\quad f(x) = \max(0, x), \overleftarrow{a}^{L}_{ij} = P_i\cdot p_{ij}
\end{align}
In our case, we start the non-linear back-propagation from the MIL layer using its sigmoidal activations for the class $i$ specific feature maps $\{p_{ij}\}$ multiplied by the pooling activation for each class $P_{i}\cdot p_{ij}$. 
Similar to Springenberg \emph{et al}.  \cite{springenberg2014striving}, we find that applying the ReLU activation function to the partial derivatives during back propagation generates Jacobian maps that are sharper and more localized to relevant objects in the input. To generate segmentation masks we threshold the sum of the Jacobian maps along the input channels. To improve the localization of cellular regions we use loopy belief propagation \cite{frey1998graphical} in an MRF to de-noise the thresholded Jacobian maps (figure~\ref{fig:jacobian_maps}).

\section{Experiments}

\subsection{Datasets}


\textbf{Breast cancer screen}: We used a benchmarking dataset of MFC-7 breast cancer cells available from the Broad Bioimage Benchmark Collection (image set BBBC021v1) \cite{ljosa2012annotated}. 
The images contain 3 channels with fluorescent markers for DNA, actin filaments, and $\beta$-tubulin at a resolution of 1024x1280. Within this dataset 103 treatments (compounds at active concentrations) have known effects on cells based on visual inspection and prior literature and can be classified into 12 distinct categories referred to as mechanism of action (MOA). We sampled 15\% of images from these 103 treatments to train and validate our model. The same proportion of the data was used to train the best model reported in \cite{ljosa2013comparison}. In total we used approximately 300 whole microscopy images during training and report test performance on all the images from the screen.

\textbf{Yeast protein localization screen}: We used a genome wide screen of protein localization in yeast \cite{Koh15042015} containing  images of 4,144 yeast strains from the yeast GFP (Green Fluorescent Protein) collection \cite{huh2003global} representing \texttildelow71\% of the yeast proteome. The images contain 2 channels, with fluorescent markers for the cytoplasm and a protein from the GFP collection at a resolution of \texttildelow1010x1335. We sampled  \texttildelow6\% of the screen and used \texttildelow2200 whole microscopy images for training. We categorized whole images of strains into 17 localization classes based on visually assigned localization annotations from a previous screen \cite{huh2003global}. These labels include proteins that were annotated to localize to more than one sub-cellular compartment. 

\subsection{Model architecture}
We designed the CNN such that an input the size of a typical cropped single cell produces output feature maps of size 1x1. The same network can be convolved across larger images of arbitrary size to produce output feature maps representing probabilities of target labels for different locations in the input image. We also aimed to show that training such a CNN end-to-end allows the model to work on vastly different datasets. We trained the model separately on both datasets while keeping the architecture and number of parameters constant. 

The basic CNN architecture includes the following layers: ave\_pool0\_3x3, conv1\_3x3x32, conv2\_3x3\_64, pool1\_3x3, conv3\_5x5\_64, pool2\_3x3, conv4\_3x3\_128, pool3\_3x3, conv5\_3x3\_128, pool4\_3x3, conv6\_1x1\_1000, conv7\_1x1\_$N_{class}$, MIL\_pool, FC\_$N_{class}$ (figure~\ref{fig:figure_1}). 
To use this architecture for MIL we use a global pooling function $g(\cdot)$ as the activation function in the MIL\_pool layer. $g(\cdot)$ transforms the output feature maps $z_i$ into a vector with a single prediction $P_i$ for each class $i$. We explore the pooling functions described above (\ref{sec:pooling_functions}). All of these pooling functions are defined for binary categories and we use them in a multi-label setting (where each output category has a separate binary target). To extend this framework we add an additional fully connected output layer to the MIL\_pool layer in order to learn relations between different categories. For the breast cancer screen this layer uses softmax activation while for the yeast data this layer uses a sigmoidal activation (since proteins can be annotated to multiple localization categories). 

\subsection{Model training}
We trained both models with a learning rate of 10\textsuperscript{-3} using the Adam optimization algorithm \cite{kingma2014adam}. We extracted slightly smaller crops of the original images to account for variability in image sizes within the screens (we used 1000x1200 for the breast cancer dataset and 1000x1300 for the yeast dataset). We normalized the images by subtracting mean and dividing by the standard deviation of each channel in our training sets. During training we cropped random 900x900 patches from the full resolution images and applied random rotations and reflections to the patches. We use the ReLU activation for the convolutional layers and apply 20\% dropout to the pooling layers and 50\% dropout to layer conv6. We trained the models within 1-2 days on a Tesla K80 GPU using \texttildelow9 Gb of memory with a batch size of 16.

\subsection{Model evaluation}
\begin{table*}[!tph]
\vskip 0.15in
\centering
\begin{tabular}{|c|ccccccc}
\cline{2-7}
\multicolumn{1}{l|}{}      & \multicolumn{4}{c|}{\bf Mean Average Precision}                                                                                 &  \multicolumn{2}{c|}{\bf Classification}                           &  \\ \cline{1-7}
\multicolumn{1}{|c|}{Model} & \multicolumn{1}{c|}{\parbox{1.7cm}{\centering cropped cell bag}} & \multicolumn{1}{c|}{\parbox{1.7cm}{\centering cropped cell manual} }& \multicolumn{1}{c|} {\parbox{1.0cm}{\centering full image}} & \multicolumn{1}{c|}{Huh} & \multicolumn{1}{c|} {\parbox{1.3cm}{\centering single loc acc.}} & \multicolumn{1}{c|} {\parbox{1.5cm}{\centering single loc mean acc.}} &  \\ \cline{1-7}
\specialcell{ Chong \emph{et al.} \cite{Koh15042015} \\ cropped cell bag }
& \specialcell{-- \\ 0.855  }
& \specialcell{-- \\ 0.742 }
& \specialcell{-- \\ --  }
& \specialcell{0.703 \\ 0.873 }
& \specialcell{ 0.935\\ 0.890  }
& \multicolumn{1}{c|}{\specialcell{ 0.808 \\ 0.799 }} \ts\\
\cline{1-7}\\
\cline{1-7}
\specialcell{ NAND $a_5$  \\NAND $a_{7.5}$\\ NAND $a_{10}$  }
& \specialcell{0.701  \\ 0.725 \\ 0.701 }
& \specialcell{0.750  \\ 0.757 \\ 0.738 }
& \specialcell{0.921  \\ 0.920 \\ \bf 0.950 }
& \specialcell{0.815  \\ 0.846 \\ \bf 0.883 }
& \specialcell{0.942 \\ \bf 0.963 \\ 0.953 }
& \multicolumn{1}{c|}{\specialcell{ 0.821 \\ 0.834 \\\bf 0.876 }} \ts\\
\cline{1-7}\\
\cline{1-7}
\specialcell{LSE $r_1$ \\ LSE $r_{2.5}$  \\ LSE $r_{5}$  \\GM $r_1$ (avg. pooling) \\ GM $r_{2.5}$ \\ GM $r_5$  \\ max pooling }
& \specialcell{0.717  \\ 0.715 \\ 0.674 \\ 0.705 \\ 0.629 \\ 0.255 \\ 0.111 }
& \specialcell{\bf 0.763 \\ 0.762 \\ 0.728 \\ 0.741 \\ 0.691 \\ 0.258 \\ 0.070 }
& \specialcell{0.925 \\ 0.925 \\ 0.933 \\ 0.915  \\ 0.888 \\ 0.405 \\ 0.125  }
& \specialcell{0.817 \\ 0.829 \\ 0.861 \\ 0.822 \\ 0.837 \\ 0.390 \\ 0.133 }
& \specialcell{0.945 \\ 0.953 \\ 0.960 \\ 0.938  \\ 0.922 \\ 0.506  \\ 0.346 }
& \multicolumn{1}{c|}{\specialcell{ 0.828 \\ 0.859 \\ 0.832 \\ 0.862 \\ 0.778 \\ 0.323 \\ 0.083 }} \ts\\
\cline{1-7}
\end{tabular}
\vskip -0.1in
\caption{ \label{table:class_yeast}
Yeast dataset results. Cropped cells indicate cells cropped from whole images based on cell coordinates from a previous segmentation pipeline. Bag and manual indicate datasets that were labeled with noisy global image labels and manually by eye, respectively. Full image indicates mean average precision on full resolution images. Huh indicates agreement with manually assigned protein localizations \cite{huh2003global}. Single loc acc. and single loc mean acc. indicate the accuracy and mean accuracy across all classes for a subset of proteins that localize to a single compartment.}
\vspace{-0.5cm}
\end{table*}

We evaluated the performance of both models at several tasks. For the yeast dataset (table~\ref{table:class_yeast}), proteins are annotated to localize to one or more sub-cellular compartments. We report the accuracy and mean classifier accuracy (across 17 classes) for a subset of 998 proteins annotated to localize to a single sub-cellular compartment in both \cite{huh2003global} and \cite{Koh15042015}. We also report the mean average precision for the all the proteins we analyzed (\texttildelow2592). For the breast cancer dataset (table~\ref{table:class_human}) we report accuracy on all of the full resolution images in the dataset and accuracy at predicting the MOA of different treatments by taking the median prediction across the 3 experimental replicates of the screen. For these predictions we use the output from the last layer of the network.

In addition to the performance on full resolution images, we evaluate the performance of the models on single cropped cells. From previous analysis pipelines using \emph{CellProfiler} we extracted centre of mass coordinates of segmented cells and used these coordinates to crop single cells from the full resolution images. We used a crop size of 64x64 for the yeast dataset and 96x96 for the breast cancer dataset. These datasets were annotated according to the labels from the full resolution images and \emph{likely include mislabeled samples}. For the yeast dataset we also report performance on \texttildelow6,300 manually labeled single cells (cropped cell manual in table~\ref{table:class_yeast}) used to train the SVM classifiers described in \cite{Koh15042015}. For these predictions we use the output from the MIL\_pool layer. 

Finally, we demonstrate that our model learns to locate regions with cells.
We generated segmentation maps identifying cellular regions in the input by back-propagating activations as described in section~\ref{sec:localization_method}. To evaluate our segmentation method we calculated the mean intersection over union (IU) between our maps and segmentation maps generated using the global otsu thresholding module in \emph{CellProfiler} which was used in \cite{Koh15042015}. We achieve a  \textbf{mean IU of 81.2\%} using this method. We found that mask pairs with low IU were mostly incorrect using Otsu thresholding. We also demonstrate that our model can generate class specific segmentation maps by back-propagating individual class specific feature maps while setting the rest of the feature maps to zero. Figure~\ref{fig:jacobian_maps} shows the Jacobian maps generated for an image with transient, cell cycle dependent protein localizations. 

\begin{table*}[!tph]
\vskip 0.15in
\centering
\begin{tabular}{|c|cccccc}
\cline{2-6}
\multicolumn{1}{l|}{}      & \multicolumn{3}{c|}{\bf All}                                                                                 &  \multicolumn{2}{c|}{\bf Held-out repetition}                           &  \\ \cline{1-6}
\multicolumn{1}{|c|}{Model} & \multicolumn{1}{c|}{\parbox{1.7cm}{\centering cropped cell bag}}  & \multicolumn{1}{c|}{full image} & \multicolumn{1}{c|}{treatment} & \multicolumn{1}{c|}{full image} & \multicolumn{1}{c|}{treatment} &  \\ \cline{1-6}
\specialcell{ Ljosa \emph{et al.} \cite{ljosa2013comparison} \\ cropped cell bag }
& \specialcell{-- \\ \bf 0.839 }
& \specialcell{-- \\ \bf 0.992 }
& \specialcell{0.94 \\ 0.99 }
& \specialcell{ -- \\ \bf 0.980  }
& \multicolumn{1}{c|}{\specialcell{ -- \\ \bf 0.984 }} \ts\\
\cline{1-6}\\
\cline{1-6}
\specialcell{ NAND $a_5$  \\NAND $a_{7.5}$\\ NAND $a_{10}$  }
& \specialcell{0.778  \\ 0.758 \\ 0.738 }
& \specialcell{0.957 \\ 0.956 \\ 0.957 }
& \specialcell{0.99  \\ 0.99 \\ 0.99 }
& \specialcell{0.915\\ 0.915 \\ 0.958 }
& \multicolumn{1}{c|}{\specialcell{ 0.957 \\ 0.957 \\  0.971 }} \ts\\
\cline{1-6}\\
\cline{1-6}
\specialcell{LSE $r_1$ \\ LSE $r_{2.5}$  \\ LSE $r_{5}$  \\GM $r_1$ (avg. pooling) \\ GM $r_{2.5}$ \\ GM $r_5$  \\ max pooling }
& \specialcell{0.816 \\ 0.807 \\ 0.750 \\ 0.803 \\ 0.750 \\ 0.382 \\ 0.185 }
& \specialcell{0.952 \\ 0.924 \\ 0.972 \\ 0.941  \\ 0.960 \\ 0.683 \\ 0.448  }
& \specialcell{0.99 \\ 0.94 \\ 0.99 \\ 0.98 \\ 0.99 \\ 0.71 \\ 0.48 }
& \specialcell{0.915 \\ 0.888 \\ 0.940 \\ 0.924  \\ .924 \\ 0.651  \\ 0.452 }
& \multicolumn{1}{c|}{\specialcell{ 0.943 \\ 0.871 \\ 0.957 \\ 0.943 \\ 0.957 \\ 0.686 \\ 0.429 }} \ts\\
\cline{1-6}
\end{tabular}
\vskip -0.1in
\caption{ \label{table:class_human}
Breast cancer dataset results. Cropped cell bag indicate cells cropped from whole images based on cell coordinates  from a previous segmentation pipeline and labeled with noisy global image labels. Full image indicates accuracy on full resolution images. Treatment indicates accuracy predicting treatment MOA by taking the median over 3 experimental repetitions. Held-out repetition indicates results leaving out any images from experimental repetitions that were sampled in the training set.}
\vspace{-0.5cm}
\end{table*}


\section{Discussion \& Conclusions}
Our proposed model links the benefits of MIL with the classification power of CNNs. We based our model on similarities between the aggregation function $g(\cdot)$ used in MIL models and pooling layers used in CNNs. This approach allows our model to learn instance and bag level classifiers for full resolution microscopy images without ever having to segment or label single cells. Our results indicate that convolutional MIL models achieve better performance across all evaluations against several benchmarks. 

The benchmarks we compare against include a classification CNN with a similar architecture trained on cropped cells given noisy whole image (bag level) labels, and a naive implementation of convolutional MIL which uses global max pooling for $g(\cdot)$. For the yeast dataset we compare with results published in \cite{Koh15042015} using an ensemble of 60 binary SVM classifiers. For the breast cancer dataset we compare with results published in \cite{ljosa2013comparison} using factor analysis. Our models outperform previously published results for both datasets with much fewer pre and post processing steps. It's clear that the naive max pooling implementation, which perfectly satisfies the standard MIL assumption, isn't suited for convolutional MIL applied to microscopy datasets.

We found that for all the convolutional MIL models, mean average precision is higher when evaluated on manually labeled cropped cells than when evaluated on cropped cells labeled with bag level labels (table~\ref{table:class_yeast}). For the model trained on cropped cells with bag level labels, we see the opposite trend (a drop in performance when evaluating on labeled cropped cells). This clearly indicates the utility of the MIL pooling layer. The models learn to identify cells in the full resolution images that correspond true phenotype categories given only the bag level annotations. This result is also shown by the class specific Jacobian maps visualized in figure~\ref{fig:jacobian_maps}. We see that different patterns in the full resolution image activate class specific output feature maps. 

For all of the bag level evaluations, we see that the NAND models perform best. We believe this can be explained by the pooling functions plotted in figure~\ref{fig:MIL_pooling_functions}. Setting the scaling factors ($a,r$) to lower values make the pooling functions approach mean of the feature maps, while for higher values the functions approach the max function. Since different phenotype categories may have vastly different densities of cells neither extreme suites all classes. The Noisy-AND pooling function accommodates this variability by learning an adaptive threshold for every class, as shown in figure~\ref{fig:MIL_pooling_functions}. 

The breast cancer screen differs from the yeast data in several ways. Human cells interact and form denser cultures while yeast are unicellular and are typically imaged at lower densities. Also, drug treatments in the breast cancer screen affect most of the cells \cite{ljosa2013comparison} while the protein localization screen contains some localization categories that only occur transiently or in a fraction of the imaged cells. For the breast cancer dataset (table~\ref{table:class_human}) we also see that that CNNs outperform the previously reported treatment accuracy \cite{ljosa2013comparison}. However, we find that the different MIL CNNs and the recognition CNN perform similarly. 
The CNNs likely perform similarly because the bag level labels correspond to almost all the cells in the images and these instances dominate the feature maps. Despite the similar performance between the MIL CNNs and the recognition CNN, the MIL CNNs have a significant advantage in that they do not require any intermediate segmentation steps. 


In summary, we found that the MIL approach we investigated offers several advantages for applications requiring classification of microscopy images. Our approach only requires a handful of labeled full resolution microscopy images. We trained the breast cancer model with only \texttildelow25 images per class. Since there are no intermediate processing steps, convolutional MIL models can be trained and tested directly on raw microscopy images in real-time. Our proposed adaptive Noisy-AND MIL pooling function performed best at classifying whole images on the challenging yeast dataset. Using the Jacobian maps we can segment cellular regions in the input image and provide predictions per cellular region as well as globally for the whole image.



{\footnotesize
\bibliography{root}}



\clearpage

\graphicspath{ {figures/}}

     

\beginsupplement
\section*{Supplementary Material}

\begin{figure}[h]
\begin{center}
\includegraphics[width=1.\textwidth]{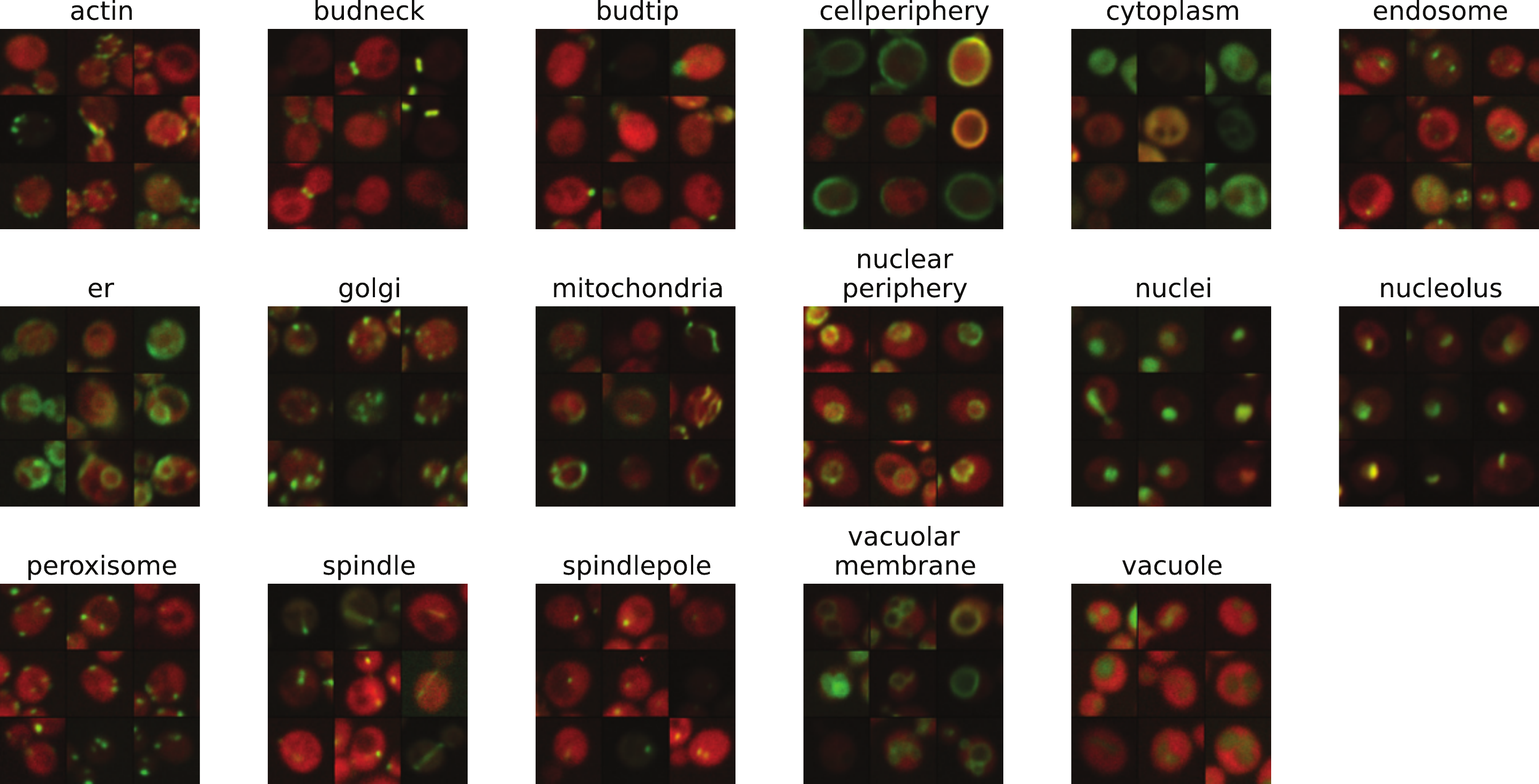}
\end{center}
\caption{Classes for yeast dataset.}
\label{fig:yeast_classes}
\end{figure}

\begin{figure}[h]
\begin{center}
\includegraphics[width=0.9\textwidth]{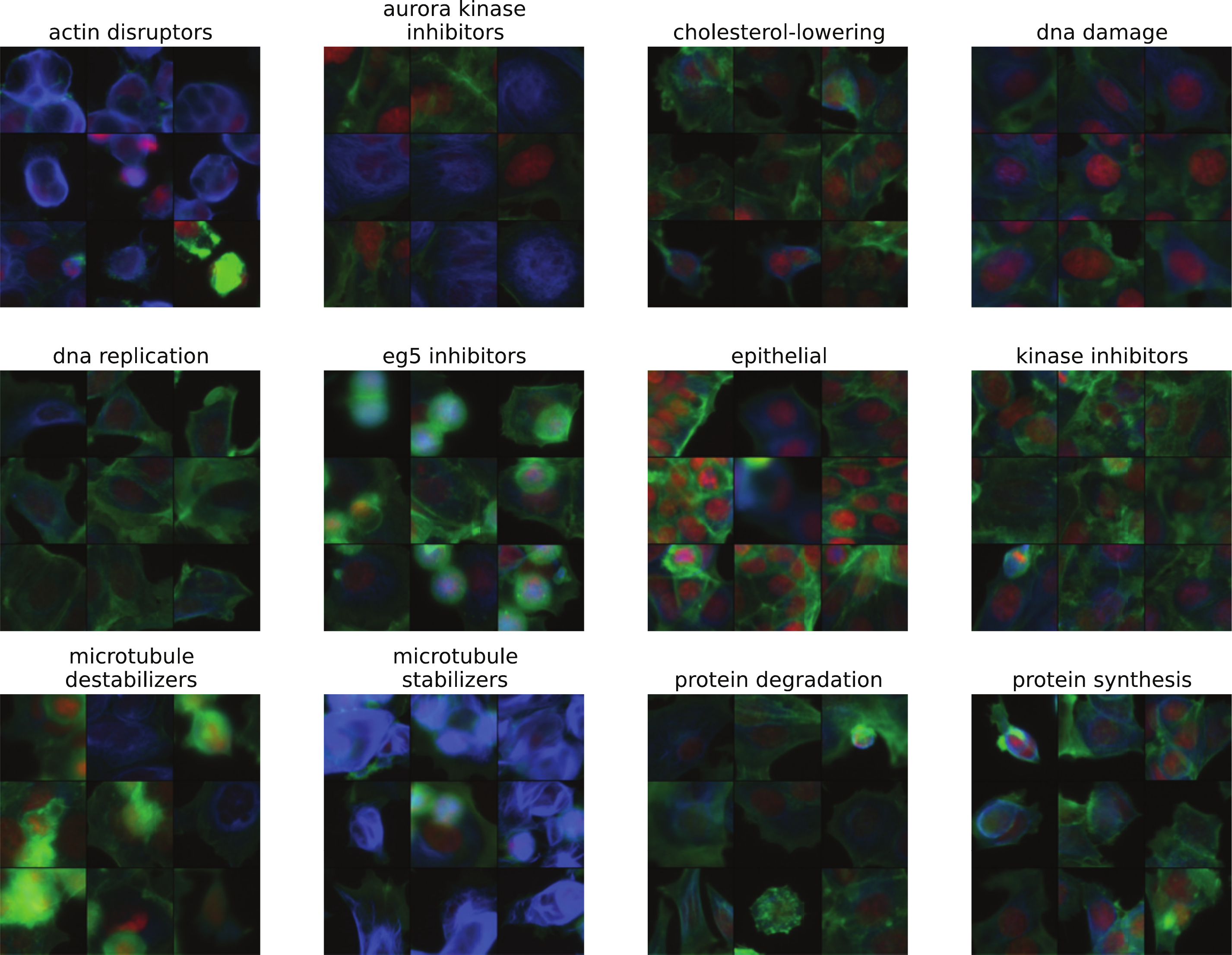}
\end{center}
\caption{Classes for breast cancer dataset.}
\label{fig:yeast_classes}
\end{figure}

\begin{figure}[h]
\begin{center}
\includegraphics[width=0.9\textwidth]{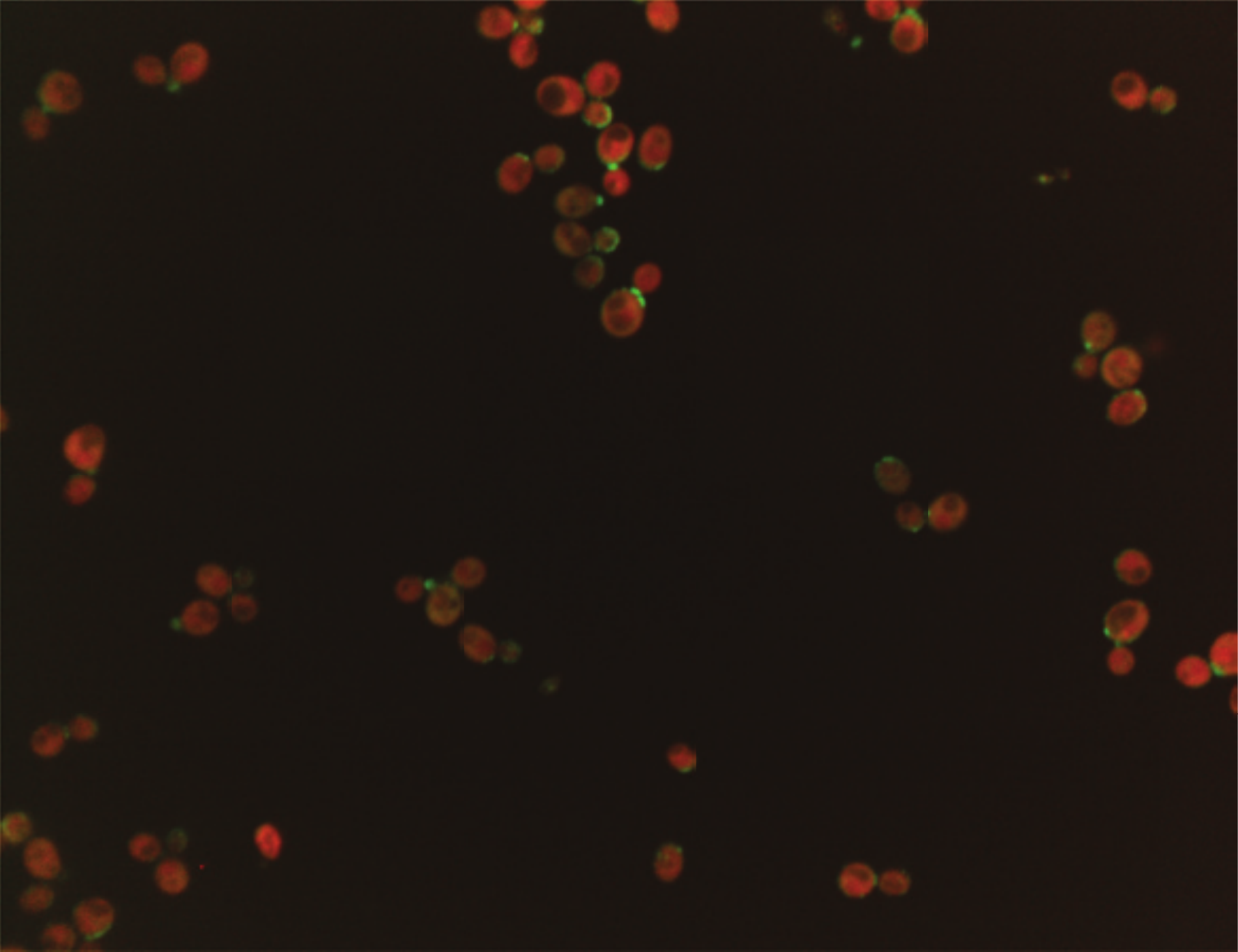}
\end{center}
\caption{Sample full resolution image for yeast dataset.}
\label{fig:yeast_classes}
\end{figure}

\begin{figure}[h]
\begin{center}
\includegraphics[width=0.9\textwidth]{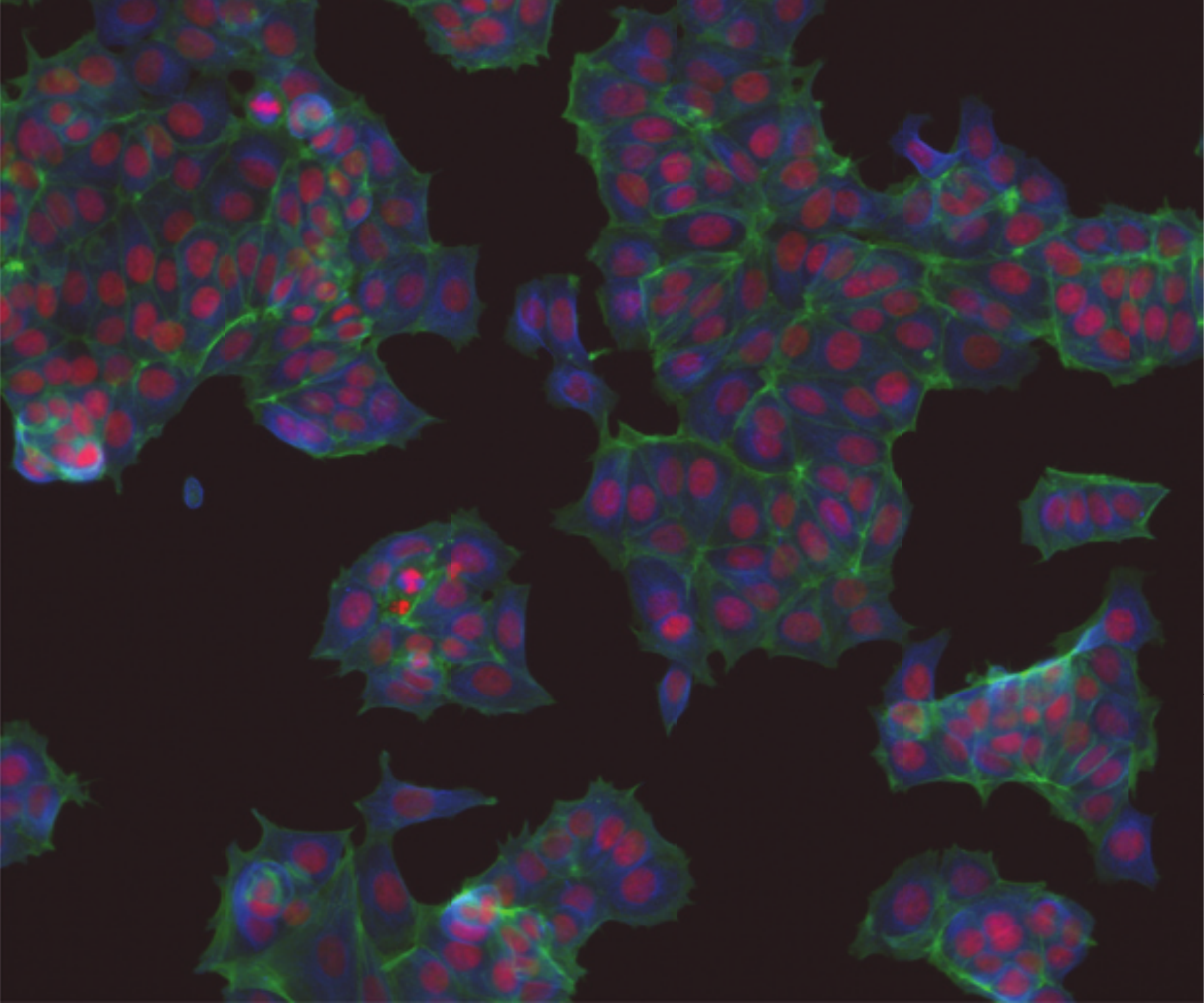}
\end{center}
\caption{Sample full resolution image for breast cancer dataset.}
\label{fig:yeast_classes}
\end{figure}


\end{document}